\documentclass[10pt,journal,letterpaper,compsoc]{IEEEtran}
\usepackage[numbers,sort]{natbib}

\ifCLASSOPTIONcompsoc
\else
\fi
\usepackage[pdftex]{graphicx}

\ifCLASSINFOpdf
\else
\fi
\usepackage{booktabs}

\usepackage{comment}

\usepackage[cmex10]{amsmath}
\usepackage{algpseudocode}

\usepackage{mdwlist}
  \usepackage[caption=false,font=normalsize,labelfont=sf,textfont=sf]{subfig}
\usepackage{url}

\begin{document}
\newcommand{\amos}[1]{#1}
%
\title{The supervised\\ hierarchical Dirichlet process}
%
%
%
%

\author{Andrew~M.~Dai and
        Amos~J.~Storkey%
\IEEEcompsocitemizethanks{\IEEEcompsocthanksitem Andrew M. Dai is with
  Google Inc. This work was completed while at the University of
  Edinburgh. E-mail: adai@google.com
\IEEEcompsocthanksitem Amos J. Storkey is with
  the Institute for Adaptive and Neural Computation, School of
  Informatics, University of Edinburgh. E-mail:
  a.storkey@ed.ac.uk}
\thanks{}}
\IEEEcompsoctitleabstractindextext{%
\begin{abstract}
We propose the supervised hierarchical Dirichlet process (sHDP), a nonparametric
generative model for the joint distribution of a group of
observations and a response variable directly associated with that whole group. We compare the sHDP with another leading
method for regression on grouped data, the supervised latent Dirichlet
allocation (sLDA) model. We evaluate our method on two real-world classification
problems and two real-world regression problems.
Bayesian nonparametric regression models based on the Dirichlet process, such as the Dirichlet process-generalised
linear models (DP-GLM) have previously been explored; these models allow flexibility in modelling
nonlinear relationships. However, until now, Hierarchical Dirichlet Process (HDP) mixtures have not
seen significant use in supervised problems with grouped data since a straightforward application of the HDP on the grouped data results in learnt clusters
that are not predictive of the responses. The sHDP solves this
problem by allowing for clusters to be learnt jointly from the group structure and from the label assigned to each group.
\end{abstract}

}

\maketitle

\IEEEdisplaynotcompsoctitleabstractindextext

%
\IEEEpeerreviewmaketitle

\section{Introduction}
\IEEEPARstart{B}{ayesian} nonparametric models allow the number of model parameters that are utilised to grow as more data is observed. In this way the
structure of the model can adapt to the data. A Dirichlet process (DP)
mixture model \citep{antoniak_dpmm} is a popular type of nonparametric model that has an
infinite number of clusters. DP mixtures are trained in an unsupervised manner
and are frequently used for problems that require model adaptation to different
data sizes or where more and more new components are likely to be represented in the data as the data size increases.

In this paper, we describe a new nonparametric \emph{supervised} model for grouped data that utilises topics, where topics
are distributions over data items that are shared across groups. We analyse the
performance of the model using experiments on both \emph{regression} and
\emph{classification} tasks. The problems of regression and
classification are ubiquitous and related; both involve labelled examples. Each example takes the form of pair consisting of a
\emph{predictor}, also known as input, covariate or independent variable, and a
\emph{response}, also known as output or dependent variable. The set of
examples is then used as data to inform models that predict the responses for test examples where the response is unknown.

Topic models such as latent Dirichlet allocation (LDA)
\citep{blei_lda} are unsupervised models of grouped data, where the \emph{topics} are distributions that are shared across the groups.
A typical example of such data is the text in the documents of a
corpus. In this context, each group is a document, and the topics are distributions over a vocabulary of terms (e.g. words). The topic distributions are shared across a number of the documents.  Each
topic can be thought of as a group of semantically related words, and inferred topics shed light on the common themes that run through
the documents. Topic modes of this form are mixed-membership models since each
document consists of a mixture of topics in different proportions. Topic models have been successful in analysing
collections of documents, including abstracts from citation databases
\citep{Griffiths2004} and newsgroup corpora. They can also be used for
a wide range of applications including data exploration, authorship
modelling \citep{rosen-zvi_author_topic_model} and information retrieval. The latent topics that are learnt are
particularly important when modelling large document collections as they can reduce the dimension of the data.

Recently, attention has turned to these models as ways of performing
regression and classification on collections of documents, where each
document possesses an associated response. The response can be
categorical, continuous, ordered or of some other type. For example, the
response could be a sentiment rating. A simple
approach to the problem of modelling document responses is to use topic models as a dimensionality
reduction method and then to regress on the resulting lower dimensional dataset. A
set of topics is learnt for the corpus using a topic model while
ignoring the document responses. Then the document responses are
regressed on the empirical topic distribution for each document. However,
this approach performs poorly in contrast to directly regressing on
the empirical word distribution for each document
\citep{Blei2010}. The topics that are learnt also often have no
relation to the responses that need to be predicted. As a result, the words that cause positive responses and those that cause negative responses end up being assigned to the same topic. This difficulty has spurred interest in supervised topic models that can learn topics that are both good models of document contents and are good predictors for document responses.

Supervised topic models (sLDA) \citep{Blei2010} are an extension of
latent Dirichlet allocation (LDA) \citep{blei_lda}. Topics that are learnt are more
useful for predicting a document's response than those obtained in
unsupervised LDA. This is because the learnt topics are oriented
around terms that are predictive of document responses. For example, in sentiment analysis tasks, the topics
learnt consist of terms that cause the document to have positive
or negative sentiment. Similarly, for financial news, the topics
consist of terms that have positive or negative effects in the
market. In contrast, unsupervised LDA learns topics that are in line with the general theme of the documents, but are often unrelated to the document responses. \citet{Blei2010} found that the predictions made by sLDA for the responses of an unseen test set were better than the predictions made using the unsupervised topics inferred by LDA. However, they found that the sLDA model only performed slightly better than LASSO regression on the empirical distribution of words for each document.

Although supervised topic models perform well, they are limited as the number of
topics in the model must be fixed in advance. This can lead to
overfitting in sLDA when there are too many topics and
regression parameters in the model so that topics are relatively
specific and do not generalise well to unseen observations. Underfitting is the opposite case
when there are too few topics and regression parameters in the
model so unrelated observations are assigned together to the same
topic. \amos{Another characteristic is the relative contribution
  of the supervised and unsupervised components to the model. In a
  fixed parametric setting it can be that one or the other (typically the unsupervised part) of these signals may dominate the likelihood, which determines the topic features. In a nonparametric setting, any dominant individual signal is captured by a set of components, leaving the remaining joint topic and supervised signal to be captured by as many additional components as are relevant.}

A number of methods can be used to choose the number of
topics, including cross-validation and model comparison techniques;
however, these are slow as the algorithm has to be restarted a
number of times and choosing the ideal number of topics from the runs
can be difficult. Bayesian nonparametric methods have emerged as a
good way to extend these models naturally to handle a flexible number
of topics.

The nonparametric supervised HDP (sHDP) model is presented in this paper. The sHDP model is a
generative supervised model that has an infinite number of topics (or
clusters) that can be used to predict a document response. The
sHDP model is a nonparametric extension of the supervised topic model
(sLDA) \citep{Blei2010}. The main contribution of the model is that it overcomes the issue of choosing
the fixed number of topics that is necessary for sLDA. The fact that the model has an infinite number
of topics also reduces the problems of underfitting and overfitting.
The sHDP can also be considered a supervised extension of the HDP mixture model described in
Section~\ref{sec:hdp}. In this paper, we show that sHDP performs better than sLDA on one dataset or comparable to sLDA with the best
performing number of topics (chosen post-hoc) on two out of three other datasets (see e.g. Figure~\ref{full-results}).

The rest of the paper is organised as follows. Section~\ref{sec:prob-desc} sets the problem and the
form of the data for the models proposed in this
paper. Then in Section~\ref{sec:chap7-existing-work}, we
briefly review some existing work on tackling the supervised learning
problem with nonparametric models and also approaches specifically for
grouped data, and goes on in Section~\ref{sec:glm} to give an introduction to generalised
linear models, and review the sLDA model (Section~\ref{sec:slda}), both of which are important in the later parts of the paper. We then introduce the supervised
HDP model in Section~\ref{sec:shdp}. Section~\ref{sec:chap7-inference}
describes the inference algorithms that are used to sample from
the posterior of the new model. Finally,
Section~\ref{sec:chap7-experiments} covers experiments with this model on real-world datasets consisting of both binary and continuous responses and compares the new model to existing models.

\section{Problem description}
\label{sec:prob-desc}
In this section we outline the structure of the problems for which this work is relevant. First, we assume that there is a set of data points divided into $D$ groups. Second, to reduce complexity, we should be willing to assume a bag of words representation can be used for each group, which amounts to assuming exchangeability among the observations within a group. Each group $i$ consists of both a variable number of data points ${x_{ij}},j=1,\dots,N_i$, which are the predictors, and a single response $y_i$. Given a set of training examples with predictors and associated responses, the task is to predict the responses on a separate test set of predictors. In the case of document modelling, $D$ is the number of documents in the corpus, each word uses one-of-$V$ encoding $x_{ij} \in \{1,\dots,V\}$  where $V$ is the size of the vocabulary of the corpus. $y_i$ is the response for the document, such as a rating or a category. In the rest of this paper, the problem and models will be described in terms of documents and words, but all the models can also be used on other kinds of grouped data.

\section{Background}
\label{sec:chap7-existing-work}
In this section we outline previous work and other methods that will be used in this paper. Many of these approaches utilise Bayesian nonparametric models to gain more flexibility than parametric models.

Due to their flexibility, there has been interest in supervised nonparametric models, such as the regression models of Gaussian processes (GPs) \citep{rasmussen_williams_book} and Bayesian regression trees. Dirichlet processes have also been adapted for supervised problems. An example of this is the Dirichlet process multinomial logit model (dpMNL) \citep{Shahbaba2007}. In this generative model, the relationship between the covariates and responses are modelled jointly using Dirchlet process mixtures. Although within each cluster the relationship is assumed to be linear, an overall nonlinear relationship occurs when the model has more than one cluster. A multinomial logit is used to model the responses conditionally on the covariates within each cluster. Thus, the regression parameters of the logit model are different for each cluster. The predicted responses are conditional on the parameters and the covariates. The dpMNL model was tested on protein fold classification, and compared with existing methods based on neural networks and support vector machines. The results showed that the dpMNL model performed significantly better.

The dpMNL has been extended to model additional response types with DP mixtures of generalised linear models (DP-GLM)
\citep{L.Hannah2011}. Whereas the dpMNL only explicitly models discrete responses, the DP-GLM can generatively model both continuous
and discrete responses using different generalised linear models. Again, the regression coefficients of the generalised linear models are different for each cluster. Priors are also placed on the coefficients, resulting in a regularised model for the response. The model was shown to have weak consistency by \citet{L.Hannah2011}, and the performance was shown to be comparable to a Gaussian process model.

Neither the dpMNL nor the DP-GLM has, to our knowledge, been applied to the problem of predicting the responses of groups of observations. The supervised topic model (sLDA) is one approach to tackling this prediction problem for grouped data (e.g. documents).
sLDA learns topics that are able to model the document responses more
accurately. The sLDA model has, however, limited flexibility since the
number of latent topics must be fixed in advance leaving it at risk of
overfitting or underfitting. There has also been work on other methods of
learning the regression coefficients or other response types such as
DMR \citep{Mimno2008}, MedLDA \citep{zhu2009medlda} and labeled LDA
\citep{ramage2009labeled}, however, these models still have a fixed
number of topics.

Hierarchical Dirichlet process (HDP) mixture models, described in
Section~\ref{sec:hdp}, are a type of Bayesian nonparametric model that
can be used instead of LDA for topic modelling. They are commonly used
as the nonparametric analog to LDA, allowing for flexible topic
modelling without being restricted to a fixed number of topics. Though
inference is more complex, Gibbs sampling and variational Bayes
techniques can still be applied. Until now, HDP mixtures have not seen
significant use in supervised problems and suffer the same problems as
unsupervised LDA in that the topics learnt are not necessarily
predictive of the responses. The sHDP model we present in this paper extends the HDP mixture model to learn topics that are good predictors of document responses.

\subsection{Hierarchical DPs}
\label{sec:hdp}
A Dirichlet process (DP) \citep{antoniak_dpmm,
  ferguson_dirichlet_process} is a stochastic process that can be
thought of as a probability distribution on the space of probability
measures. The name of the process accurately describes the fact that the DP
results in finite-dimensional Dirichlet marginal distributions,
similar to the Gaussian process that has Gaussian distributed
finite-dimensional marginal distributions. DPs are commonly used as a
prior on the space of probability measures, which give wider support
and so improved flexibility over using traditional parametric families
as priors. In addition, DPs also have tractable posteriors so making them
important in Bayesian nonparametric problems. A DP is defined in terms
of a base measure and a concentration parameter. Each draw from the DP is
itself a measure. Since there is a positive probability of
drawing a previously drawn value, the draws are discrete with
probability $1$. This makes them very useful for clustering in DP mixtures.

The HDP \citep{Teh_Jordan_HDPs} is a hierarchical extension to DPs. The hierarchical structure provides an elegant way of sharing parameters. This process defines a set of probability measures $G_i$ for $D$ pre-specified groups of data and a global probability measure $G_0$. The global measure is distributed as
\begin{equation}
G_0|\gamma,H \sim \operatorname{DP}(\gamma,H)
\end{equation}
where $H$ is the base probability measure and $\gamma$ is the concentration parameter.

The random measures for each group $i$ are conditionally independent given the global measure
\begin{equation}
G_i|\alpha_0,G_0 \sim \operatorname{DP}(\alpha_0, G_0)
\end{equation}
where $\alpha_0$ is a concentration parameter. The distribution $G_0$
varies around $H$ by an amount controlled by $\gamma$ and the
distribution $G_i$ in group $i$ varies around $G_0$ by an amount
controlled by $\alpha_0$. This can be seen as adding another level of
smoothing on top of DP mixture models. Let $\theta_{i1},\theta_{i2},\dotsc$ be i.i.d. variables distributed to $G_i$ and each of these variables is a parameter that corresponds to an observation $x_{ij}$, the likelihood of these observations being
\begin{align}
\theta_{ij}|G_i &\sim G_i \\
x_{ij}|\theta_{ij} & \sim F(\theta_{ij})
\end{align}
where $F(\theta_{ij})$ is the distribution of $x_{ij}$ given $\theta_{ij}$. This prior results in a DP being associated with each group in the model where the DPs are conditionally independent given their parent and the parameters drawn in the parent node are shared among the descendant groups. This structure can be extended to multiple levels.

The HDP requires that the data be in a pre-defined nested structure. The HDP model has been used in information retrieval tasks and used in relation with traditional TF-IDF measures \citep{cowans_ir_hdp} for measuring the score of documents in relation to a query. There are variants of HDP that model topics for documents where there is no predefined hierarchical structure (see e.g. \citep{wei_nonparametric_pachinko}).

\subsubsection{Similarity to LDA}
\label{sec:hdp_lda}
With the appropriate base measure, the HDP can be thought of as the
infinite analogue of LDA. In the HDP, the base probability measure
allows for a countably infinite number of multinomial draws and so an
infinite number of topics. This allows the number of topics to grow or
shrink according to the data. This solves the problem of finding the
best number of topics in LDA and reduces the problems of overfitting
or underfitting due to a fixed number of topics.

\subsection{Generalised linear models}
\label{sec:glm}
Often when a response is not an unconstrained continuous variable, it
is transformed into one and a normal linear model is used for
it. However, this may not always be appropriate. A \emph{generalised
  linear model} (GLM) \citep{mccullagh_glm} expands the flexibility of linear regression by
being capable of analysing data where either there may not be a linear
relation between the covariates $x$ and the response $y$ or where a
Gaussian assumption for $y$ is inappropriate. \amos{Given parameters $\boldsymbol{\eta}$, and covariates $\mathbf{x}$, a generalised linear
model is specified by a linear predictor which we denote in this
section by $\rho=\boldsymbol{\eta}^T\mathbf{x}$}, a link function $g(\cdot)$ that relates the
linear predictor to the mean $\mu$ of the response $\mu=g^{-1}(\rho)$ and
a probability distribution from the exponential family that gives the distribution of
the response $y$ with mean $\operatorname{E}(y|\cdot)=\mu$. In this
paper, we only consider canonical link functions though others can be
used when needed. \amos{The canonical link function is a choice of
  link function such that $\rho$ is the natural parameter in the
  exponential family distribution.} The distribution of the response
may also be an exponential dispersion family that has an additional
dispersion parameter denoted as $\delta$. We denote this as
ExpFam$(\mu, \delta)$. The generalised linear model for response $y$ takes the form
\begin{align}
  p(y|\rho,\delta)=h(y,\delta)\exp \left\{ \frac{\rho y-A(\rho)}{\delta}\right\},
\end{align}
where $A(\rho)$ the log-normaliser.

Different forms of responses can be modelled using different choices of $h$ and $A$. In particular, there is a Gaussian distribution on $y$,
\begin{align}
  p(y|\rho,\delta)=\frac{1}{\sqrt{2 \pi \delta}}\exp\left\{-\frac{1}{2 \delta} (y-\rho)^2\right\}
\end{align}
when $h(y,\delta)=\left(1/ \sqrt{2 \pi \delta}\right)\mathrm{e}^{-y^2/2}$ and $A(\rho)=\rho^2/2$. This is a normal linear model with a mean of $\rho$ and variance of $\delta$.

When $y$ is binary, a binomial distribution can be used with the number of trials $n=1$, so that $y$ is distributed as
\begin{align}
  p(y|\rho)=\rho^y(1-\rho)^{1-y}
\end{align}
which uses the canonical logit link function
$g(\rho)=\ln(\rho/(1-\rho))$ and the binomial distribution for
$y$. This choice of distribution and link function results in a
logistic regression model.

\subsection{The supervised topic model}
\label{sec:slda}
The supervised topic model (sLDA) \citep{Blei2010} is an extension of LDA to supervised problems. It partially overcomes the problem that the topics that are learnt cannot be controlled in the LDA model. The learnt topics in LDA act to reduce the dimension of the data but may not be predictive of a document's response as they will correspond to the general themes of the corpus. sLDA overcomes this problem by jointly learning topics and their regression coefficients for the document responses. The response for a document is predicted by averaging over the empirical topic allocations for a document.

The generative process for each document $i$ is the following. Let $K$ be the fixed number of topics, $N_i$ the number of words in document $i$, $\boldsymbol{\phi}_{1:K}$ the topics where each $\boldsymbol{\phi}$ is a distribution over the vocabulary, $\alpha$ a parameter for topic proportions, and $\boldsymbol{\eta}$ and $\delta$ the response parameters.

\begin{enumerate}
\item Draw topic proportions $\boldsymbol{\vartheta}_i \sim \textrm{Dirichlet}(\alpha)$.
\item For each word (enumerated by $j$)
  \begin{enumerate}
  \item Draw a topic assignment $z_{ij} \sim \textrm{Multinomial}(\boldsymbol{\vartheta}_i)$.
  \item Draw a word $w_{ij}|z_{ij} \sim \textrm{Multinomial}(\boldsymbol{\phi}_{z_{ij}})$.
  \end{enumerate}
\item Draw the document response $y |
  z_{i,1:N_i},\boldsymbol{\eta},\sigma^2 \sim
  \textrm{ExpFam}(g^{-1}(\boldsymbol{\eta}^{\top}\bar{\mathbf{z}}_i),\delta)$ where $\bar{\mathbf{z}}_i=1/N_i \sum_{j=1}^{N_i} z_{ij}$.
\end{enumerate}

This implements a GLM for the document responses: ExpFam is a distribution from the exponential family, $g$ is the link
function and $\delta$ is the
dispersion parameter for the distribution. The linear predictor in the GLM model for the response is $\boldsymbol{\eta}^{\top}\bar{\mathbf{z}}$ where $\bar{\mathbf{z}}$ are the empirical frequencies of the topics in the document and $\boldsymbol{\eta}$ are the regression coefficients. Exchangeability of the topic assignments imply posterior parameter symmetries in the GLM model, were a full Bayesian solution obtained. However if we are constrained to a MAP inferential setting there is no possibility of parameter symmetry and so there is symmetry breaking of the exchangeability of topic assignments. The inference process used must be sensitive to this broken symmetry. An inference process that considers the generation process for the document contents first enables consistent topic labels to be determined. Then the document's response is chosen conditional on those contents, and hence on topic labels that have a consistent meaning. An alternative to this is to choose a model where $y$ is regressed on the topic proportions for the document, $\boldsymbol{\vartheta}$. However, this may result in some topics being estimated that just explain the response variables while other topics only explain the document words.

In the sLDA model, the parameters $\alpha$, $\boldsymbol{\phi}_{1:K}$,
$\boldsymbol{\eta}$ and $\delta$ are treated as constants to be
estimated. Approximate maximum-likelihood estimation is then performed
with a variational expectation-maximisation (EM) method, similar to
that for LDA. Collapsed Gibbs sampling can also be used for inferring
the topics jointly as in LDA.

The models we propose in this paper solve the issue sLDA has of requiring the number of topics to be fixed from the start. This can result in overfitting or underfitting if the number of topics is unsuitable for the dataset. Though the
number of topics can be chosen based on a training set, the process
can be difficult and time
consuming.

\section{The supervised HDP (sHDP) model}
\label{sec:shdp}

The supervised HDP (sHDP) model proposed in this paper can automatically learn the necessary number of topics to model the responses of documents on training data. It is a Bayesian nonparametric model so that a potentially infinite number of latent clusters can be used for prediction. The sHDP model extends the HDP mixture model to learn clusters that align with document responses. The relationship between the data points and the responses is modelled with a generalised linear model on the clusters to which the data points in a document have been allocated. A regression coefficient is associated with each cluster, and the document's response is regressed on the mean of these coefficients.

In the sHDP model, unlike sLDA, the number of topics does not need to
be fixed in advance. This is beneficial in supervised problems since
it is unclear how many latent topics will be necessary to model the
data and the response conditional on the document. The response is
modelled by a generalised linear model (GLM) conditioned on the topics
that have been assigned to the observations in the document. Since the
number of instantiated topics can vary and each topic has a regression
coefficient, the number of instantiated regression coefficients also
varies given the current number of instantiated topics. In the
generative process, a regression coefficient is sampled for each topic
in addition to sampling a distribution over the vocabulary. In effect,
a product base measure is used for the topics where one component is a
prior over the vocabulary and the other is a prior for the regression
coefficient. This treats the regression coefficients as random
variables, whereas in sLDA, the regression coefficients are treated as
constants. This modelling of the regression coefficients results in a
regularised regression model for the response variables. Each topic
can also be assigned a vector of regression coefficients for categorical responses.

The model is thus
\begin{align}
  G_0 \sim & \textrm{DP}(\gamma H) \\
  G_i \sim & \textrm{DP}(\alpha G_0)\\
  \boldsymbol{\theta}_{ij} = (\theta_{ij}^X, \theta_{ij}^Y) \sim & G_i\\
  x_{ij}|\theta_{ij}^X \sim & f(\theta_{ij}^X)\\
  y_i|\boldsymbol{\theta}_{i\cdot}^Y \sim & \textrm{ExpFam}(g^{-1}(\overline{\boldsymbol{\theta}_{i\cdot}^Y}), \delta)
\end{align}
where $\overline{\boldsymbol{\theta}_{i\cdot}^Y} = (1/N_i) \sum_j
\theta_{ij}^Y$ is the linear predictor for the GLM, $g$ is its link
function and $\delta$ is the dispersion parameter for the exponential family distribution. $i$ ranges over each document, $j$ ranges over each observation in that document, $\gamma$ denotes the concentration parameter for the corpus-level DP and $\alpha$ denotes the concentration parameter for the document-level DP. The base measure $H=H^Y \times H^X$ consists of a measure for the regression parameters $\theta^Y \sim H^Y$ and another measure for the topic parameters $\theta^X \sim H^X$. $G_0$ is the corpus-level random measure that acts as the base measure for the document-level random measure $G_i$.

\begin{figure}
  \centering
  \includegraphics[scale=0.6]{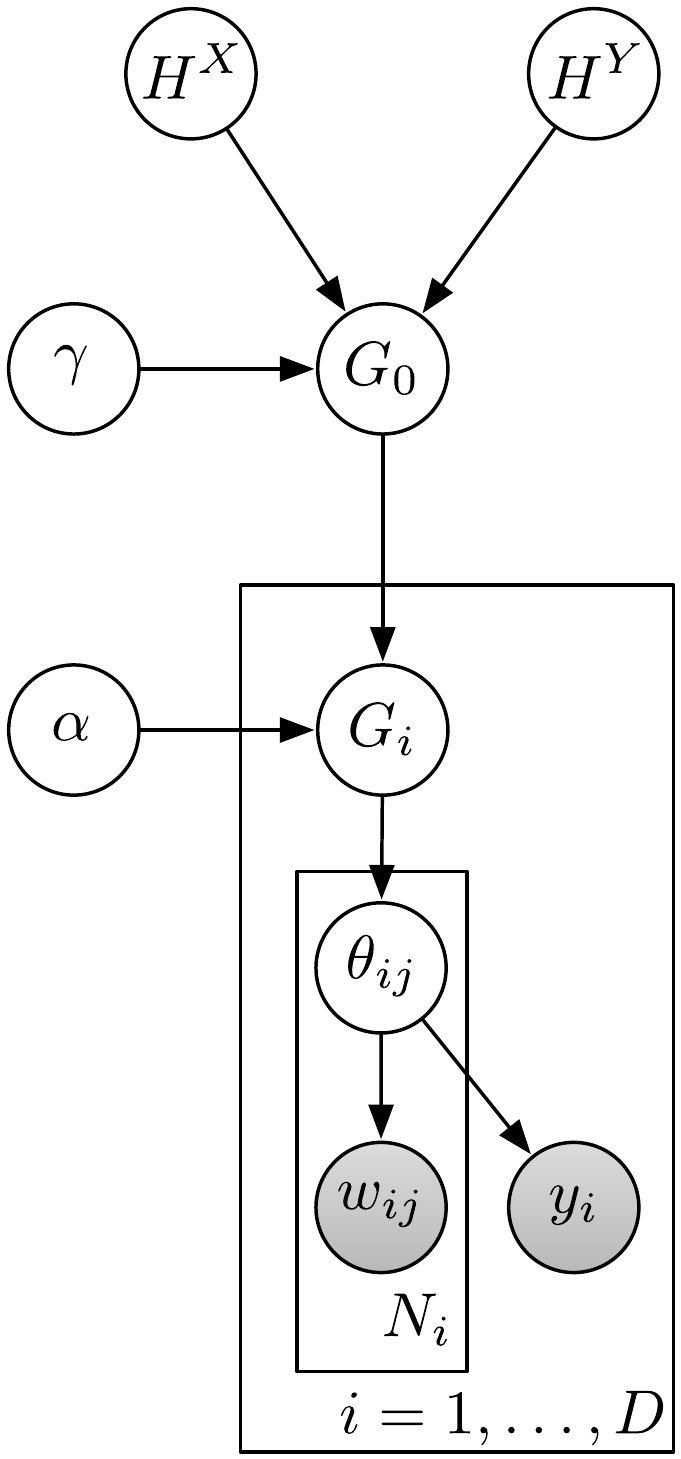}
  \caption[Supervised HDP graphical model]{The supervised HDP model
    where the observed variables are the words $w_{ij}$ denoting word
    $j$ in document $i$ and the document label $y_i$.}
      \label{sup-graphic-model-hdp}
\end{figure}

Due to the clustering property of the DP, some data points will share
the same parameters $\boldsymbol{\theta}$, which can be represented as those data
points being assigned to the same topic. The prior density for the
regression parameters is typically $H^Y=\textrm{N}(0,\zeta)$. For
topic modelling, the documents consist of words, and the prior density
for the cluster parameters is $H^X=\textrm{Dirichlet}(\alpha^w)$,
where $\alpha^w$ is the parameter for a symmetric Dirichlet
distribution. $f$ is the likelihood of $\theta^X$ given the
observations $x$. In a topic modelling problem,
$f(\theta^X)=\textrm{Multinomial}(\cdot|\theta^X)$. When coupled with
its conjugate prior, the Dirichlet distribution, the topic parameters
$\theta^X$ can be integrated out, allowing for collapsed Gibbs
inference to be performed by just keeping track of the word to topic
allocations and the regression coefficients for the topics. The GLM
model for the responses allows the responses to be continuous,
ordinal, categorical and other types depending on the form of the
GLM. If the base measure for the coefficients $H^Y$ is chosen to be
Gaussian, the maximum a posteriori (MAP) solution for the coefficients is similar to the solution for $L_2$ penalised regression. A graphical model is shown in Figure~\ref{sup-graphic-model-hdp}.

The generative process for the full model is:
\begin{enumerate}
\item Draw (from their prior distributions) the concentration parameters for the global DPs $\gamma$. Likewise, draw the concentration parameters for the lower-level DPs $\alpha$ from their priors.
\item Draw a global distribution over topics and their regression coefficients $T_0 \sim \textrm{DP}(\gamma, H)$.
\item Now for each document $i$,
  \begin{enumerate}
  \item Draw a distribution over topics $T_i \sim \textrm{DP}(\alpha, T_0)$.
  \item For each word $w_{\mathit{ij}}$,
    \begin{enumerate}
    \item Draw a topic $(\boldsymbol{\theta}^X_{\mathit{ij}}, \theta^Y_{\mathit{ij}}) \sim T_{i}$.
    \item Draw a word $w \sim \textrm{Multinomial}(\boldsymbol{\theta}^X_{\mathit{ij}})$.
    \end{enumerate}
  \item Draw a response for the document $y \sim
\textrm{ExpFam}(g^{-1}(\overline{\boldsymbol{\theta}^Y_{i \cdot}}),\delta)$ where $\overline{\boldsymbol{\theta}^Y_{i \cdot}}=(1/N_i) \sum_j \theta^Y_{ij}$.
  \end{enumerate}
\end{enumerate}

The sHDP learns topics that both model document contents well and are
predictive of document responses without the need to choose a fixed
number of topics beforehand. This structured approach to supervision
allows the model to be easily extended to incorporate additional
information from documents to aid in predicting the response such as
the authors of a document or the research group which authored a
document, which can be inferred through the grouped author-topic model
\citep{Dai2011a}. For example, there is the problem of predicting the venue
where a paper is published by learning the venues where the research
group has previously published. Another example could be the problem
of predicting a set of keywords or categories for a paper by learning
which categories have previously been picked by the research group for
those topics. Allowing for the topics to be supervised can also
give more control over the types of topics that are learnt by the
sHDP in case the unsupervised topics are not interesting for a
particular task. Finally, the sHDP model allows for unlabelled data to be used as part of the training set in semi-supervised problems. This allows supervised topics to be learnt that take into account the content of unlabelled documents so that the topics can better model the entire corpus instead of just the labelled documents.

\section{Inference}
\label{sec:chap7-inference}
Since posterior inference is intractable in DP-based models, approximations must be used. Collapsed Gibbs sampling is the most
common technique used to sample from the posteriors of these models,
and it can also be applied to the model described in this
paper. For topic modelling problems, the
Dirichlet base measure for each topic, which is a distribution over
the vocabulary, is conjugate to the multinomial
likelihood for the words. This enables the topic parameters to be integrated out. Thus at each
iteration and based on the Chinese restaurant process, collapsed Gibbs
sampling can be used to sample the topic allocations. The regression
coefficients can then be sampled
from their posteriors in some cases such as for a Gaussian response
and can be approximated in other cases. The following
sections describe inference in the proposed sHDP model.

\subsection{The sHDP model}
Since the base measure for the topic regression coefficients will not
in general be conjugate to the response model, the non-conjugate
auxiliary variable sampling algorithm (alg. 8) described by
\citet{neal_markov_chain_dirichlet_process_mixture} is used to sample
the topic allocations, $z_{ij}$ where $z_{ij}=k$ indicates that word $w_{ij}$ is allocated to topic $k$.
The main difference from inference for the HDP
mixture model is in sampling the topic allocation variable and the
topic regression coefficients. The conditional distribution for the
topic allocation has an additional term for the conditional likelihood
of the topic parameters given the document response. Gibbs sampling
proceeds as below.
\begin{enumerate}
\item For each document $i$,
\begin{enumerate}
\item Let $n_{ik}$ denote the number of words in document $i$ allocated to topic $k$, and let a superscript $-ij$ for a variable denote the terms excluding the $ij$th term. For each word $w_{ij}$, sample the topic allocation $z_{ij}$ using
\suspend{enumerate}
\begin{equation}
\label{eq:samp-shdp-topic}
\begin{split}p(&z_{ij}=k|\mathbf{z}^{-ij},w_{ij},\boldsymbol{\beta})\\
&\propto\begin{cases}(n_{i k}^{-ij}+\alpha\beta_k)f^{-w_{ij}}_k(w_{ij}) p(y_i|\mathbf{z}^{-ij},z_{ij}=k,\boldsymbol{\eta}), \\ \mbox{if }k=z_{i'j'}\mbox{ for some } (i',j') \neq (i, j)\\
\alpha\beta_{\mathrm{new}} f_{\text{new}}(w_{ij}) p(y_i|\mathbf{z}^{-ij},z_{ij}=k,\boldsymbol{\eta^{\text{new}}}), \\ \mbox{if
}k=k^{\textrm{new}}\end{cases}\end{split}\end{equation}
where $\boldsymbol{\eta}^{\text{new}}=(\boldsymbol{\eta}, \eta^{k^{\text{new}}})$, $\eta^{k^{\text{new}}} \sim \textrm{N}(0,\zeta)$, $f_k$ is the distribution of the word given the other words allocated to topic $k$ and $f_{\text{new}}$ is the probability of the word in an empty topic.

If a new topic $k^{\mathrm{new}}$ is sampled during one of the steps above, then draw $b \sim \textrm{Beta}(1,\gamma)$, set the new weight $\beta_{k^{\mathrm{new}}}=b \beta_{\text{new}}$ and set the new $\beta_{\text{new}}$ to $(1-b)\beta_{\text{new}}$. The value $b$ corresponds to the weight of the new atom that is instantiated from the Dirichlet process. Also, set $\boldsymbol{\eta}$ to the value of $\boldsymbol{\eta}^{\text{new}}$.

\resume{enumerate}
\item Sample $m_{ik}$, where $k$ ranges over the topics, by generating $n_{i k}$ uniformly distributed random variables $u_1,\dots,u_{n_{i k}}$ between 0 and 1 and setting
\begin{align}
\label{eq:samp-shdp-m}
 m_{ik}=\sum_{m=1}^{n_{i k}} \mathbf{1}\left[u_m \geq \frac{\tau \beta_k }{ \tau \beta_k + m}\right]
\end{align}
where $\mathbf{1}$ is the indicator function.
\end{enumerate}

\item Sample $\boldsymbol{\beta}$
  from \begin{align} \label{eq:samp-shdp-beta}
    (\beta_1,\dotsc,\beta_K,\beta_{\text{new}}) \sim
    \textrm{Dirichlet}(m_{\cdot 1},\dotsc,m_{\cdot K},\gamma) \end{align}.
\end{enumerate}

For a continuous response assuming $\gamma=1$,
\begin{align}
  p(y_d|\mathbf{z},\boldsymbol{\eta}) \propto \exp(-(y_d-\boldsymbol{\eta}^{\top}\bar{\mathbf{z}})^2)
\end{align}
and for a binomial response where $y_d \in \{0, 1\}$,
\begin{align}
  p(y_d|\mathbf{z},\boldsymbol{\eta}) = (\boldsymbol{\eta}^{\top}\bar{\mathbf{z}})^{y_d}(1-\boldsymbol{\eta}^{\top}\bar{\mathbf{z}})^{1-y_d}.
\end{align}

During prediction, the posterior of $\bar{\mathbf{z}}$ is needed over the test documents. This is calculated by removing the terms that depend on the response $y$ from the conditional distributions so that inference on the test documents is identical to unsupervised sHDP. The posterior for the test samples can be sampled by replacing \eqref{eq:samp-shdp-topic} with
\begin{equation}
  \label{eq:samp-shdp-pred-topics}
\begin{split}p(&z_{ij}=k|\mathbf{z}^{-ij},w_{ij},\boldsymbol{\beta})
  \propto \\
  &\begin{cases}(n_{i k}^{-ij}+\alpha\beta_k)f^{-w_{ij}}_k(w_{ij}), \\
    \mbox{if }k=z_{i'j'}\mbox{ for some } (i',j') \neq (i, j)\\
\alpha\beta_{\mathrm{new}} f_{\text{new}}(w_{ij}) , \mbox{if
}k=k^{\textrm{new}}\end{cases}\end{split}\end{equation}
and sampling the allocations and counts for the test documents.

\subsection{Parameter posteriors and prediction}

The topic regression coefficients are sampled after each round of
sampling the topic assignments. We also performed experiments where
the topic assignments were sampled for several rounds in between
sampling the regression coefficients but this made little difference to prediction performance. The topic coefficients can be updated for sHDP by regressing only on the topics that are allocated to at least one observation. We will describe cases for a Gaussian and binary response in this section, though other models for the response can be used too.

\subsubsection{Gaussian model}

In the Gaussian model, we place a Gaussian prior on the regression
coefficients. The model response can be rewritten as
\begin{align}
  \mathbf{y}=\mathbf{X}\boldsymbol{\eta}+\mathbf{c}
\end{align}
where $\mathbf{y}$ is a length-$D$ vector of document responses,
$\mathbf{X}$ is a $D \times \infty$ matrix of cluster to document
allocation counts, $\boldsymbol{\eta}$ is a vector of regression
parameters for each topic and $\mathbf{c}$ are the residuals. Let
$\mathbf{X}$ be the matrix where row $d$ is the empirical topic
distribution for document $d$. Since only a finite number of topics
have non-zero counts in the corpus, the columns in $\mathbf{X}$ that
have zero counts and their corresponding $\boldsymbol{\eta}$ entries can
be ignored, so making posterior computation tractable.

The posterior distribution for the parameters $\boldsymbol{\eta}$ is then
a Gaussian distribution.
\begin{align} \label{gaussianPosterior}
  \boldsymbol{\eta} \sim \textrm{Gaussian} \left( \left( \mathbf{X}^\top
    \mathbf{X}+\zeta \mathbf{I} \right)^{-1} \mathbf{X}^\top
  \mathbf{y}, \left( \mathbf{X}^\top
    \mathbf{X}+\zeta \mathbf{I} \right)^{-1} \right)
\end{align}
where $\zeta$ is the prior variance for the concentration parameters and \textbf{I} denotes the identity matrix.

For prediction, topics are sampled for test documents as in \eqref{eq:samp-shdp-pred-topics}. The empirical topic distribution is sampled over a number of iterations with any topics that are instantiated or any topics that are removed during this period ignored. The remaining empirical topic distributions for each document are averaged and used to calculate the expectation of the response.

For the sHDP model, this is calculated as
\begin{align}
  \textrm{E}[\mathbf{y}|\mathbf{z},\boldsymbol{\eta}] \approx \boldsymbol{\eta}^{\top}\textrm{E}[\bar{\mathbf{z}}].
\end{align}

\subsubsection{Binomial model}

For the logistic regression GLM model, the likelihood is
\begin{align}
  l(\boldsymbol{\eta}) \propto -\sum_{d=1}^D \log(1+\exp(-y_d \boldsymbol{\eta}^{\top}\bar{\mathbf{z}}_d))-\frac{\zeta}{2}\boldsymbol{\eta}^{\top}\boldsymbol{\eta}.
\end{align}

The gradient is then
\begin{align}
\label{eq:binomial-grad}
  \nabla_{\boldsymbol{\eta}} l(\boldsymbol{\eta})=\sum_d(1-\sigma(y_d\boldsymbol{\eta}^{\top}\bar{\mathbf{z}}_d))y_d\bar{\mathbf{z}}_d-\zeta \boldsymbol{\eta}
\end{align}
where $\sigma(\cdot)$ is the logistic sigmoid function,
\begin{align}
  \sigma(x)=\frac{1}{1+\exp(-x)}.
\end{align}

We place a Gaussian prior distribution on the regression coefficients,
however since there is no conjugate prior, the posterior distribution
is not available in closed form. To sample from the exact posterior
for the coefficients, the
Gibbs sampling
method presented by \citet{Groenewald_bayesian_logistic_regression} and
used in topic models by \citet{mimno2008gibbs} can be used. However,
we found that this method took numerous iterations to converge for a
given topic assignment due to the high number of coefficients. As a
result, in our results, we instead sample from an approximation to the
posterior. A common approximation to use is the
Laplace
approximation, which involves sampling from a Gaussian centred at the MAP
estimate of the parameters with a covariance matrix that is the Hessian of the
unnormalized log posterior.
The limited-memory BFGS
algorithm can be used to find the MAP estimate of the parameters
\citep{Minka2003}.

For prediction, topics are sampled for test documents as in
\eqref{eq:samp-shdp-pred-topics}.

For the sHDP model, the distribution of the response is given by
\begin{align}
  p(y_d=1|\mathbf{z},\boldsymbol{\eta}) \approx \frac{\exp(\boldsymbol{\eta}^{\top}\textrm{E}[\bar{\mathbf{z}}])}{1+\exp(\boldsymbol{\eta}^{\top}\textrm{E}[\bar{\mathbf{z}}])}
\end{align}

\amos{For simplicity, we also consider using the MAP estimate of the
  parameters directly. In many cases we find there is not a
  significant performance benefit of using parameter sampling over
  using the MAP solution directly.}

A sampling step of the supervised HDP algorithm that samples the
regression coefficients (the sampled model) is shown below in
pseudocode. To initialize, words are randomly allocated to topics so
$\mathbf{z}$ is set randomly, $K$ is set to the maximum value of
$\mathbf{z}$, $\mathbf{m}$ and $\boldsymbol{\beta}$ are sampled from
\eqref{eq:samp-shdp-m} and \eqref{eq:samp-shdp-beta} respectively.

\begin{algorithmic}
\State \textbf{Input}: Corpus with $D$ documents, a response for each
document $y_i$ and $N_i$ words in
document $i$. Old model parameters $\mathbf{z}, \mathbf{m},
\boldsymbol{\eta} \textrm{ and } \boldsymbol{\beta}$.
\State \textbf{Output}: New model parameters: $\mathbf{z}, \mathbf{m},
\boldsymbol{\eta} \textrm{ and } \boldsymbol{\beta}$.
\For{$i\gets 1, D$}
\For{$j\gets 1, N_i$}
\State Sample $z_{ij}$ from \eqref{eq:samp-shdp-topic}
\EndFor
\For{$k\gets 1, K$}
\State Sample $m_{ik}$ from \eqref{eq:samp-shdp-m}
\EndFor
\EndFor
\State Sample $\boldsymbol{\beta}$ from \eqref{eq:samp-shdp-beta}.
\If{$\mathbf{y}$ is distributed as a Gaussian}
\State Sample the regression parameters $\boldsymbol{\eta}$ from
\eqref{gaussianPosterior}.
\EndIf
\If{$\mathbf{y}$ is distributed as a binomial}
\State Find the MAP value of the regression parameters $\boldsymbol{\eta}^M$ by using
L-BFGS to solve \eqref{eq:binomial-grad}.
\State Sample $\boldsymbol{\eta} \sim N(\boldsymbol{\eta}^M,\mathbf{H})$ where $\mathbf{H}$ is the Hessian of the
unnormalized log posterior.
\EndIf
\end{algorithmic}

\section{Experiments}
\label{sec:chap7-experiments}
We conducted experiments on four real-world datasets. First, we
considered the classification problem of determining the effect of financial newswires on the direction of change of the closing prices of a set
of stocks. Second, we focused on the classification problem of
determining whether movie
review sentences are
positive or negative. Third, we addressed the regression problem of
predicting a rating for a full movie review and fourth, the regression
problem of predicting the popularity of a document. The datasets were
preprocessed to keep the terms with the highest total TF-IDF
score. TF-IDF is a measure of how important a term is for a document
in a corpus. The score is calculated as $\textrm{tf}(w) \times \log
D/n_w$ where tf is the frequency of the term $w$ in the document, $D$ is the number of documents and $n_w$ is the number of documents where the term $w$ occurs. This is summed across all the documents for each term, and the highest scoring terms are kept.

The newswire classification dataset consists of a set of real-world newswires
extracted from \emph{Reuters} about the stocks in the S\&P 500 on
different days over a year up to May 2011. The newswires were labelled
with the companies that were mentioned in the wire. These labels were
used so that only newswires whose stocks on days that had more than an
8\% positive change or 3\% negative change from the previous day were
considered. These cutoffs were chosen so that the number of declining
stocks were similar to the number of rising ones, and to ignore minor
changes of prices due to other factors. This resulted in a dataset of
1,518 documents and a vocabulary of 1,895.

The review snippet \emph{classification} dataset \citep{Pang2005} consists of reviews from the \emph{Rotten Tomatoes} website with reviews that were marked as fresh labelled as positive reviews and reviews that were marked as rotten labelled as negative reviews. The dataset contains 5,331 positive snippets with the same number of negative ones and a vocabulary of 4,310.

The review snippet \emph{regression} dataset \citep{Pang2005} consists of
reviews written by four film critics where the writer additionally
assigned a rating to his or her review. The ratings were normalised to
be between 0 and 1. Any terms that appeared in more than 25\% of the
documents were removed as were any terms that appeared fewer than 5
times. Only the remaining top 2,179 terms by TF-IDF score were then
kept. The ratings for each document were preprocessed to normalise the
scores by applying a log transform. There was a total of 5,005
documents.

The document popularity regression dataset is a dataset of submission
descriptions from the \emph{Digg} website with the associated number
of votes that each submission received. The number of votes were again
normalised by applying a log transform. The dataset consisted of
a vocabulary of 4,120 across 3,880 documents.

Experiments were performed with the sHDP model and the sLDA
model. Both models were implemented using MCMC methods (collapsed
Gibbs sampling in the case of sLDA with the \citet{chang_lda_r} implementation) and predictions were done using an
equivalent sample in both instances. \amos{For sLDA we also applied a
  variational approach with the \citet{chong_slda} implementation, and the results for sLDA are given for both collapsed Gibbs and variational inference
  approaches.} For sLDA, for collapsed Gibbs sampling, 3000 iterations were used and for variational inference, EM was ran until the
relative change in the likelihood bound was less than 0.01\%.

The accuracy for classification problems and predictive $R^2$ for regression problems after five-fold cross-validation were calculated. Predictive $R^2$ is defined as
\begin{align}
  pR^2=1-\frac{\sum_d (\hat{y_d}-y_d)^2}{\sum_d (y_d-\bar{y})^2},
\end{align}
where $y_d$ are the observed responses, with $d$ ranging over the
documents, $\hat{y_d}$ is the response predicted by the model and
$\bar{y}=1/D \sum_{d=1}^D y_d$ is the mean of the observed
responses. This value gives the proportion of variability in the data
set that is accounted for by the model and is often used to evaluate
the goodness of fit of a model. A value of $1.0$ is obtained when the
regression line perfectly fits the data. We present accuracy and predictive $R^2$ results that are calculated
on the full set of predictions. We also give indicators of the minimum and maximum difference in performance (across the folds) of each method, relative to the sampled HDP.

In the experiments, the prior standard deviation of the parameters
$\zeta$ was tested with three values ($1$, $5$ and $10$) on each fold's
training set by splitting the fold's training set into a smaller training and
validation set and choosing the best value on the validation set. This was
also done when choosing the prior standard deviation for sLDA. $\alpha^w$ for the sHDP model
was set to $0.01$ on datasets
similar to previous experiments with HDP. In the sHDP, the standard prior Gamma($1, 1$) was placed on $\alpha$ and $\gamma$ and these are sampled during inference.
For sHDP, learning took place over $2,000$ iterations with the
coefficients being sampled every iteration. For predicting the
responses of the test documents, $500$ iterations of topic
sampling were used to allow the inferred topics to converge. The
number of iterations was chosen by looking at the trace plots of the
residuals and the regression coefficients, which appeared to converge by that number of iterations. To
compare our models, we carried out experiments using sLDA with
variable numbers of topics so that performance with sLDA with the best
performing number of
topics on the test set can also be compared.

We show results for a sHDP inference algorithm
that uses
the MAP estimate of the regression coefficients (which reduces
computation time) during inference and uses a fixed set of coefficients
at test time. We also show results for an algorithm that samples from
the posterior of those coefficients during training and
test time. For the
Gaussian model, we sample from the posterior as in Eq. \eqref{gaussianPosterior}. Using a Gibbs sampling
method to sample the coefficients of the binomial model took many
iterations to converge so we sampled the coefficients from a Laplace
approximation. Finally, we also do experiments with a
2-step algorithm in which unsupervised topics for the documents are
first learnt as in a HDP model and then a GLM model is trained on top of the learnt
topics to predict document labels. In this way, the performance of
jointly training the topics and the GLM model can be compared with training
the two in separate steps.

\subsection{Results}

\begin{figure*}
  \centering
  \subfloat[Newswire]{\includegraphics[scale=0.7]{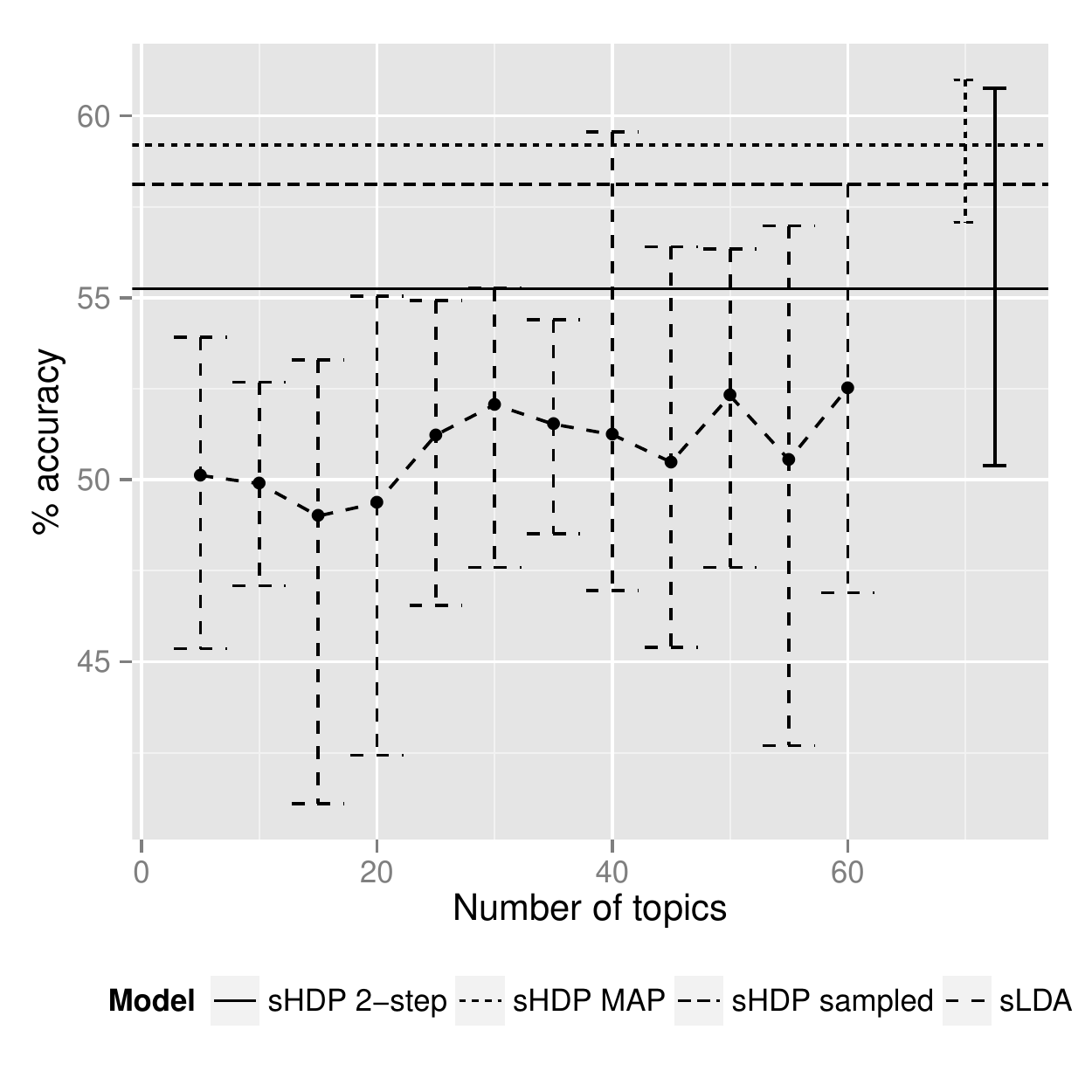}
      \label{reuters-results}}
    \hfil
    \subfloat[Movie Snippet Classification]{\includegraphics[scale=0.7]{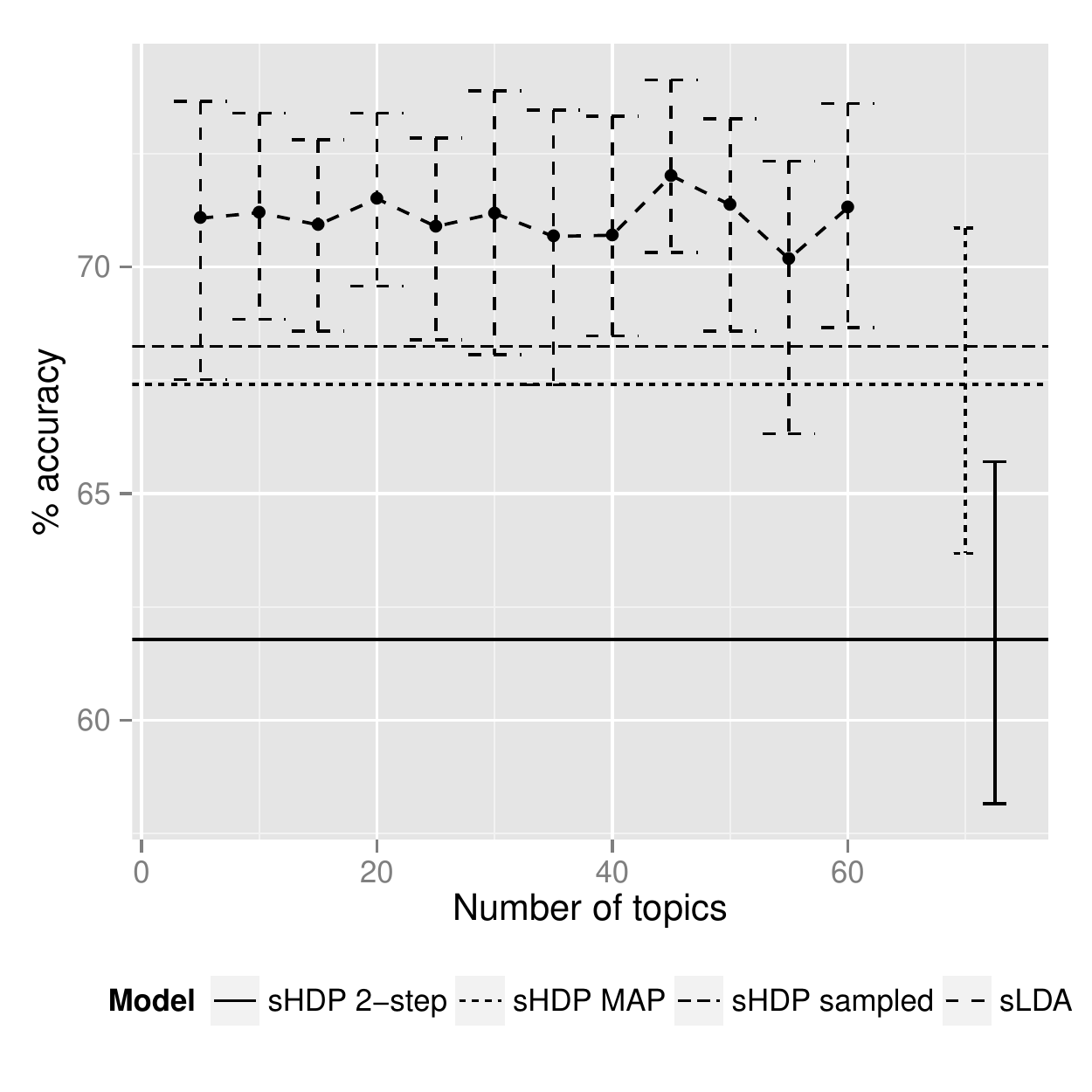}
      \label{polarity-results}}
    \hfil
    \subfloat[Movie Reviews]{\includegraphics[scale=0.7]{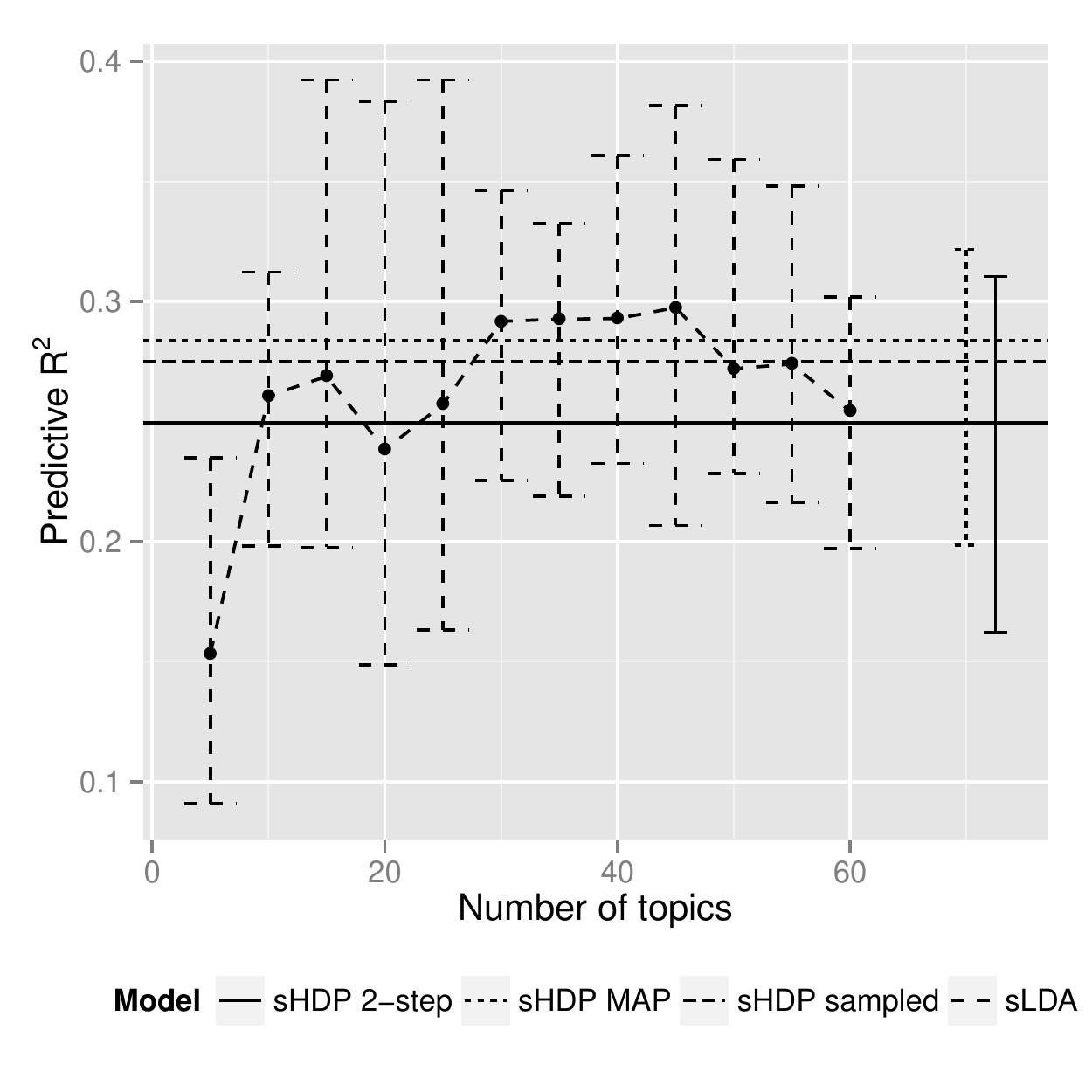}
      \label{movies-results}}
    \hfil
    \subfloat[Document Popularity]{\includegraphics[scale=0.7]{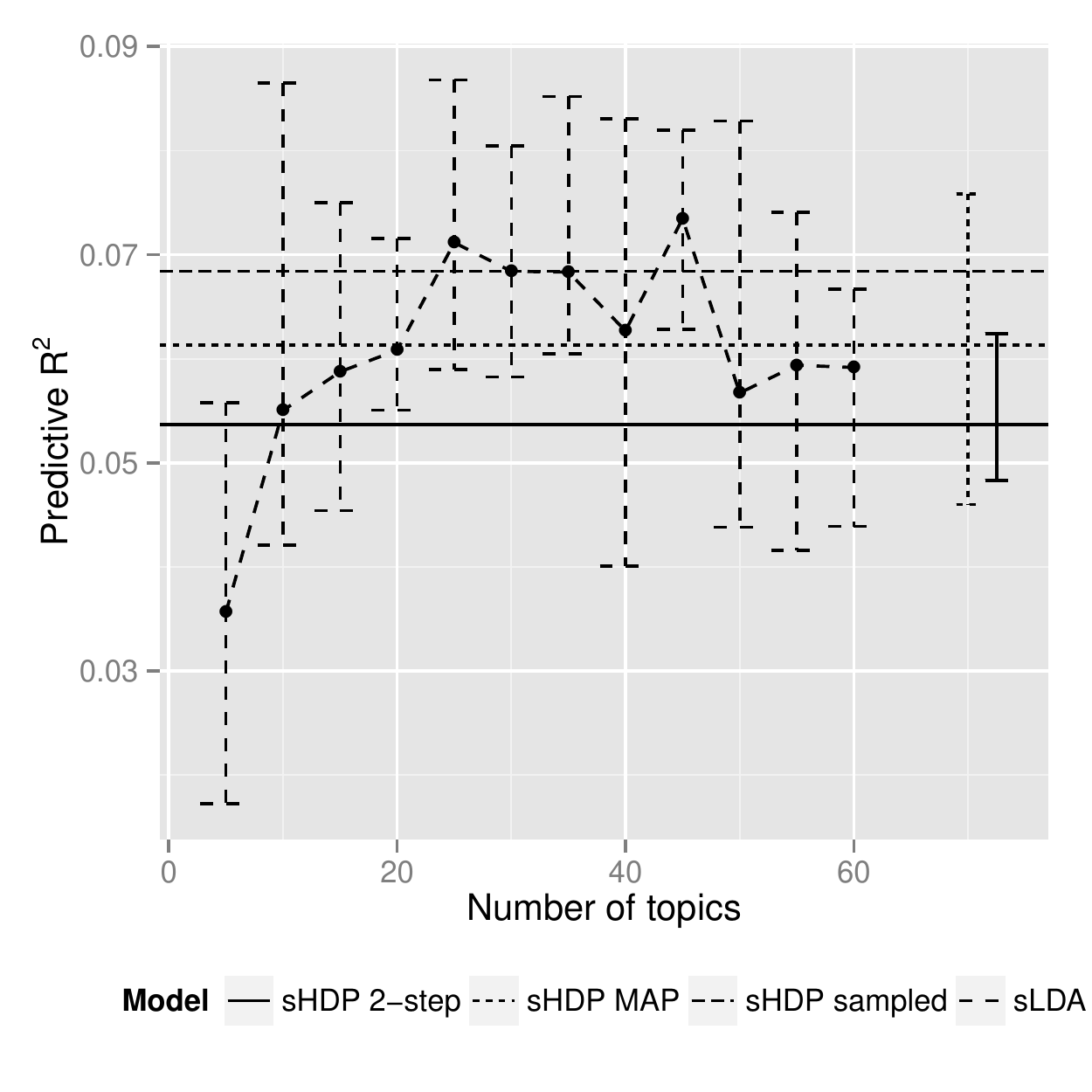}
      \label{digg-results}}

  \caption{Results for the test datasets after 5-fold
    cross-validation. Classification results are given for (a) the newswire dataset and (b)
    the movie snippet dataset.
    Regression results ($R^2$) for the entire dataset are given for (c) the movie reviews dataset and (d) the document popularity dataset.
    sLDA Gibbs performance is shown for each number of
    topics. Variational EM performed as well as or worse than Gibbs
    sampling and is omitted for space. For sHDP, the
    performance with MAP parameters, with parameters sampled from their
    posteriors and with a 2-step supervised approach where an unsupervised HDP model is learnt and a GLM model
  trained on top of that is shown. The upper and lower bars show the minimum and maximum performance of each method \emph{relative} to the performance of the sampled sHDP (minimum and maximum taken over the 5 folds). This allows the reader to see whether a method performs better or worse than sampled sHDP across all the folds.}
\label{full-results}
\end{figure*}

Figure~\ref{full-results} shows that the supervised
HDP (sHDP) model performs significantly better than the sLDA model on
the newswire dataset. For almost all models, sLDA inference using
Gibbs sampling performs better than with variational EM, so we have not shown the variational EM to avoid clutter. sHDP performs competitively against sLDA with the
best performing number of topics (as chosen on the test set) on the
remaining datasets except for the movie snippet dataset. In the movie
snippet dataset, sLDA outperformed the sHDP across the number of
topics. From the
results it can be seen that for sLDA, picking the right number of topics is key to getting
good performance. Moreover, picking too few topics or too many in some
cases can cause big drops in performance. On the other hand, for the
sHDP, the model yields good performance without having to pick the
number of topics. For sHDP, the results also show that
sampling the regression coefficients from their posteriors make little
difference to the results compared to using the MAP value of the
coefficients. Additionally, the simple alternative of learning the topics
unsupervised in a HDP model and then training a GLM model on top (a
2-step supervised approach) performs significantly worse than jointly learning
the topics with the sHDP model.

\begin{figure*}
  \centering
  \subfloat{\includegraphics[scale=0.7]{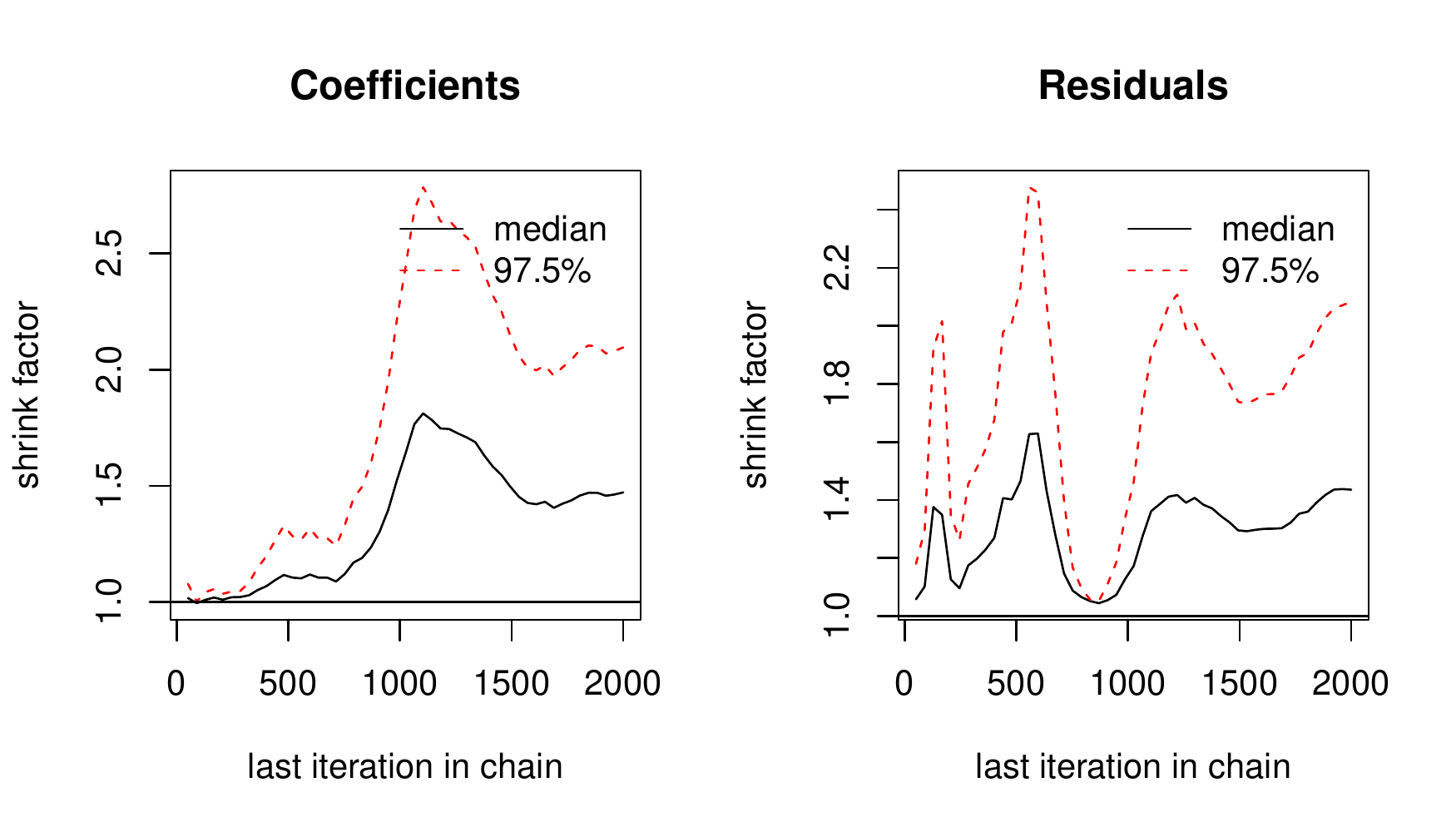}
     }
    \hfil
    \subfloat{\includegraphics[scale=0.7]{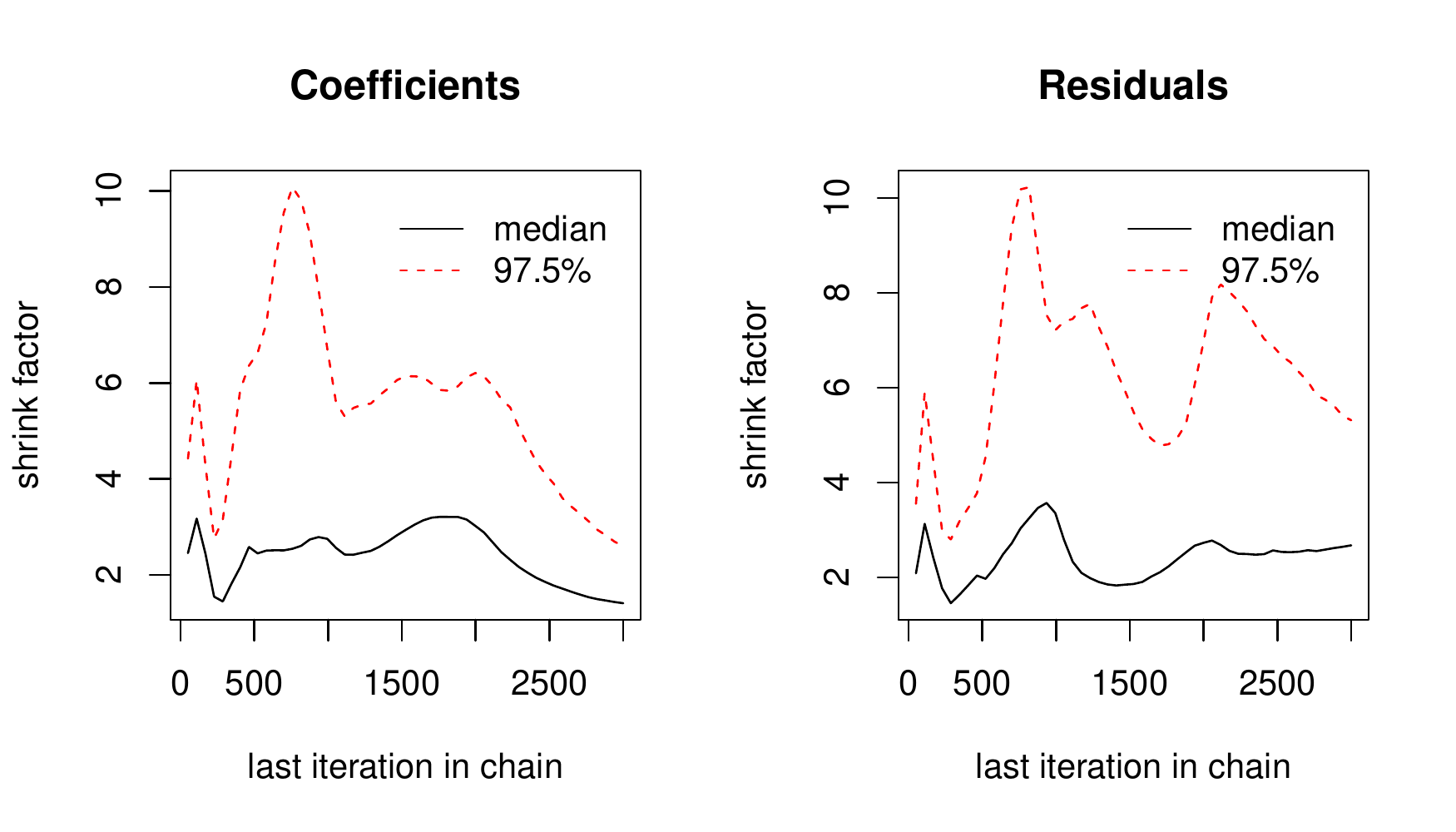}
      }
  \caption{Chain convergence for (TOP) the sHDP model and
  (BOTTOM) the Gibbs sampled sLDA model with 40 topics. Gelman-Rubin-Brooks plots show how Gelman and Rubin's
    shrink factor for the L2 norm of the regression coefficients
    and the L2 norm of the residuals changes across iterations
    during inference for the movie regression dataset. This is shown for 4 parallel MCMC
    chains with different starting values. Values close to 1 indicate
  convergence. From these graphs it can be seen that the Gibbs sampled
sLDA model is slower to mix compared to the sHDP model.}
\label{gelman-plot}
\end{figure*}

The better
performance of sHDP compared to sLDA for the newswire dataset and
competitive performance with the other datasets is partly due to the increased
flexibility of the model and better mixing during inference as can be
seen in Figure~\ref{gelman-plot}. The
increased flexibility comes from the model having an infinite number
of topics to model the documents and responses. The better mixing
results from the fact that during inference, clusters can be
instantiated or unneeded ones can be removed while sampling. Since newly instantiated clusters are empty, it is easier for words to change topic and be allocated to a new cluster. In contrast, in sLDA each topic almost always has a significant number of words allocated to it, making it difficult for the distribution of a topic to change. This has the effect of smoothing over term contributions for each topic. Thus, the fact that there are more specific topics in the sHDP model helps to improve performance.

From the relatively low accuracy scores and large
standard deviations, it can be seen that the labels on the newswire
dataset are much harder to predict than on the movie review
dataset. The standard deviations for the newswire scores imply that
the data is much more noisy since newswires only indirectly influence
stock movements. In addition, only closing stock prices are used, which means that it is possible there were changes in the
stock price from the general movement of the industry or the
market. However, the sHDP is able to pick out these subtle signals
whereas sLDA with both types of inference algorithms was unable to.

Figure~\ref{gelman-plot} is a Gelman-Rubin-Brooks plot
\citep{brooks_gelman} which shows how Gelman and Rubin's shrink factor \citep{Gelman1992}
changes as the number of iterations changes during inference. The plot of the shrink
factor is calculated from 4 parallel MCMC chains from different starting
points. We present results for the Gaussian sHDP model and the Gibbs sampled sLDA
model with 40 topics. The shrink factor is calculated by
comparing the within-chain and between-chain variances for each
variable of interest. The factor predicts that the chains have converged if the
output from the chains are indistinguishable, which is given by the
factor approaching 1. In the plots, the shrink factors for the L2 norm of
the regression coefficients and the L2 norm of the
residuals for the two models are shown. As can be seen from the plots, the shrink factor for the
sLDA model is significantly higher than that of the sHDP model,
indicating the sHDP model exhibits better mixing.

We also conducted experiments by regressing directly on
the empirical word distribution for each document with L1 regularized
generalized linear models and the GLMNET R package. The regularization
parameter was chosen through cross validation on the training set of
each fold.  The accuracies are $61\%$ for newswire dataset, $75\%$ for the movie
snippet dataset,  and $R^2$ of $0.44$ for the
movie rating dataset and $0.064$ for the document popularity dataset. Hence this approach marginally outperformed both sLDA
and sHDP on newswire, movie snippet and movie review datasets, but it is outperformed by the sHDP on the document popularity dataset. Given the large number of available parameters, the model flexibility provides benefits for the L1 GLM, but makes it particularly sensitive to particular word usage. It is therefore understandable that this model does best on the three datasets with more coherent word usage patterns, but is less powerful than a sampled sHDP on document popularity, where the word usage within a topic is more varied. The extra topic structure also enables an understanding of the topic-related dependencies which we expand on in the next section.

\subsection{Analysis of strong topics and terms}

For the sHDP model, the top positive and negative topics, in terms of
their regression coefficients and their most frequent terms for the
movie review problem, are shown in
Table~\ref{inferred-topics-review}. The topics do not generally
correspond to themes such as film genre or style. Instead of this, the
topics contain many names such as actors and directors. This is
because the flexibility of a nonparametric model means that the top
positive and negative topics consist of very few terms and are
allocated to actors and directors that are consistently reviewed well
or poorly. This flexibility results in strong topics that are grouped
around consistently performing actors or directors but the topics are
less coherent since they are associated with so few documents. Topics
that consist of more terms, even if those are strong terms, generally
have smaller regression coefficients since the effect of the different
terms is averaged over other words in the same topic. Strong terms are
spread among the top positive and negative topics, for example,
positive topic 5 contains the positive term \emph{charming} and
negative topic 2 contains many negative terms such as
\emph{unfortunately}, \emph{worse} and \emph{problem}. Since many of
the topics have actor and director names such as \emph{Tom Hanks} in
positive topic 2 and the \emph{Coen brothers}: \emph{Ethan} and
\emph{Joel} in positive topic 4, it can be seen that specific actors
and directors are associated with consistently better or poorer movie
review scores.

The terms in the top topics for sHDP seem to correspond to people's names, such as \emph{cameron} and \emph{miller}, with only one topic focusing on terms that intuitively should have a strong contribution to a movie rating. This shows that the topics being learnt are divided into those that correspond to the content of the corpus and those that are more focused on general terms that affect the rating of a movie.
The most positive topic and negative topics
have no association with film genres and are more concentrated on
specific actors, which are more likely to perform consistently.

The top positive and negative topics for the 2-step
algorithm in which a GLM model is trained on top of unsupervised topics
learnt with a HDP model are shown in
Table~\ref{inferred-topics-2step-review}. These topics are different
in that there are no topics like
negative topic 2 from the sHDP model and in
general there are much fewer sentiment-related terms in the strong
topics. The fact that the regression coefficients are smaller in
magnitude also indicates that sentiment or rating-predictive terms are
more spread out among the topics meaning that individual topics are
less predictive of the rating.

\begin{table}
  \centering
  \caption[Strongest topics for the movie review dataset using sHDP]{The most positive and negative learnt topics, in terms of their regression coefficients, from the movie review regression dataset with sHDP.}
  \label{inferred-topics-review}
  \begin{tabular}{cccc}
    \toprule
  \begin{tabular}{c}
    + Topic 1 (8.3)
    \\ \midrule
    jeff\\
    philip\\
    lane\\
    write \\
    miller\\
    party\\
    cameron\\
    kate\\
    bus\\
    instead \\
  \end{tabular}
&
  \begin{tabular}{c}
    + Topic 2 (6.5)
    \\ \midrule
    tom\\
    hanks\\
    roth\\
    tim \\
    store\\
    eric\\
    speed\\
    rob\\
    wallace\\
    appear \\
  \end{tabular}
&
  \begin{tabular}{c}
    + Topic 3 (6.1)
    \\ \midrule
    philip\\
    calls\\
    happiness\\
    features \\
    baker\\
    helen\\
    jane\\
    human\\
    hoffman\\
    feelings \\
  \end{tabular}
\\ \midrule
  \begin{tabular}{c}
    + Topic 4 (6.1)
    \\ \midrule
    ethan\\
    brothers\\
    joel\\
    journey \\
    singing\\
    constant\\
    blake\\
    process\\
    wonderfully\\
    george \\
  \end{tabular}
&
  \begin{tabular}{c}
    + Topic 5 (5.8)
    \\ \midrule
    six\\
    aaron\\
    neil\\
    howard \\
    matt\\
    company\\
    teacher\\
    nature\\
    charming\\
    buddy \\ 
  \end{tabular}
\\ \midrule
  \begin{tabular}{c}
    $-$ Topic 1 (-6.2)
    \\ \midrule
    beneath\\
    child\\
    series\\
    son\\
    someone\\
    kills\\
    flaws\\
    winner\\
    onto\\
    record \\
  \end{tabular}
&
  \begin{tabular}{c}
    $-$ Topic 2 (-6.0)
    \\ \midrule
    that's\\
    least\\
    supposed\\
    watching\\
    unfortunately\\
    lot\\
    flat\\
    worse\\
    pretty\\
    problem \\
  \end{tabular}
&
  \begin{tabular}{c}
    $-$ Topic 3 (-5.4)
    \\ \midrule
    dogs\\
    aaron\\
    score\\
    animal\\
    golden\\
    air\\
    martin\\
    ball\\
    dog\\
    charles \\ 
  \end{tabular}
\\ \midrule
  \begin{tabular}{c}
    $-$ Topic 4 (-4.1)
    \\ \midrule
    rachel\\
    breaking\\
    anthony\\
    ten\\
    harry\\
    warner\\
    thinks\\
    quinn\\
    strikes\\
    dog \\ 
  \end{tabular}
&
  \begin{tabular}{c}
    $-$ Topic 5 (-4.0)
    \\ \midrule
    nelson\\
    daughter\\
    roger\\
    christopher\\
    con\\
    bergman\\
    travolta\\
    leslie\\
    flashback\\
    simon \\ 
  \end{tabular}
&
\\ \bottomrule
\end{tabular}
\end{table}

\begin{table}
  \centering
  \caption[Strongest topics for the movie review dataset using a
  2-step supervision algorithm where the supervised GLM model is learnt on
  top of a set of unsupervised topics from the HDP.]{Strongest topics
    for the movie review regression dataset using the alternative
  2-step approach where a supervised GLM model is learnt on
  top of a set of unsupervised topics from the HDP.}
  \label{inferred-topics-2step-review}
  \begin{tabular}{cccc}
    \toprule
  \begin{tabular}{c}
    + Topic 1 (6.0)
    \\ \midrule
    carter\\
    appeal\\
    happiness\\
    passion \\
    beat\\
    handsome\\
    walk\\
    easy\\
    political\\
    british \\
  \end{tabular}
&
  \begin{tabular}{c}
    + Topic 2 (4.9)
    \\ \midrule
    brothers\\
    brother\\
    ethan\\
    blake \\
    constant\\
    joel\\
    nelson\\
    bank\\
    wonderfully\\
    doing \\
  \end{tabular}
&
  \begin{tabular}{c}
    + Topic 3 (4.8)
    \\ \midrule
    taylor\\
    jim\\
    alexander\\
    serious \\
    jessica\\
    matthew\\
    shock\\
    usual\\
    meaning\\
    year's \\
  \end{tabular}
\\ \midrule
  \begin{tabular}{c}
    $-$ Topic 1 (-5.5)
    \\ \midrule
    bridges\\
    faces\\
    jeff\\
    wants\\
    sister\\
    rose\\
    baker\\
    gregory\\
    university\\
    cinematographer\\
  \end{tabular}
&
  \begin{tabular}{c}
    $-$ Topic 2 (-5.3)
    \\ \midrule
    donald\\
    shadow\\
    sutherland\\
    professor\\
    conspiracy\\
    reporter\\
    linda\\
    charlie\\
    bobby\\
    amanda \\
  \end{tabular}
&
  \begin{tabular}{c}
    $-$ Topic 3 (-5.3)
    \\ \midrule
    patricia\\
    india\\
    charlotted\\
    jungle\\
    bruce\\
    current\\
    cat\\
    fish\\
    russian\\
    studio \\ 
  \end{tabular}
&
\\ \bottomrule
\end{tabular}
\end{table}

The top positive and negative topics for the newswire dataset and
their most frequent terms are given in
Table~\ref{inferred-topics-reuters-shdp}. These topics are more
cohesive than those for the movie review dataset. The top positive
topic contains very strong positive terms such as \emph{higher},
\emph{strong}, \emph{rise} and \emph{record}, which all imply good
stock performance. The top negative topic also contains strongly
negative terms such as \emph{cut}, \emph{fall}, \emph{decline} and
\emph{drop}, which clearly indicate bad performance. Similarly to the
top topics for the movie review dataset, it can be seen that some
industries consistently have better or poorer stock performance. For
example, negative topic 2 consists of companies such as
\emph{prudential} and \emph{metlife} along with terms such as
\emph{insurers} and \emph{insurance}. The negative coefficients
indicate that the insurance industry may be performing badly. Positive topic 2 with terms such as \emph{defense}, \emph{military} and \emph{shareholders} indicates that companies involved with the military and defence are associated with rising stock prices.

\begin{table}
  \centering
  \caption[Strongest topics for the newswires dataset using sHDP]{The most positive and negative learnt topics, in terms of regression coefficients, from the newswires dataset with sHDP.}
  \label{inferred-topics-reuters-shdp}
  \begin{tabular}{cccc}
    \toprule
  \begin{tabular}{c}
    + Topic 1 (20.5)
    \\ \midrule
    prices\\
    higher\\
    since\\
    high \\
    level\\
    strong\\
    rise\\
    above\\
    record\\
    hit \\
  \end{tabular}
  &
  \begin{tabular}{c}
    + Topic 2 (15.9)
    \\ \midrule
    itt\\
    defense\\
    publicly\\
    water \\
    shareholders\\
    segments\\
    military\\
    aerospace\\
    announcement\\
    holds \\
  \end{tabular}
  &
  \begin{tabular}{c}
    + Topic 3 (15.4)
    \\ \midrule
    fedex\\
    raised\\
    outlook\\
    june \\
    monday\\
    bellwether\\
    pace\\
    housing\\
    foreign\\
    whole \\
  \end{tabular}
\\ \midrule
  \begin{tabular}{c}
    + Topic 4 (15.1)
    \\ \midrule
    north\\
    america\\
    american\\
    second \\
    latin\\
    six\\
    europe\\
    china\\
    avon\\
    september \\
  \end{tabular}
&
  \begin{tabular}{c}
    + Topic 5 (14.7)
    \\ \midrule
    wells\\
    fargo\\
    friday\\
    effect \\
    between\\
    larger\\
    repurchase\\
    accounting\\
    scheduled\\
    bonds \\ 
  \end{tabular}
\\ \midrule
  \begin{tabular}{c}
    $-$ Topic 1 (-16.4)
    \\ \midrule
    cut\\
    fall\\
    declines\\
    third-quarter\\
    ended\\
    volume\\
    third\\
    drop\\
    decline\\
    tuesday \\
  \end{tabular}
&
  \begin{tabular}{c}
    $-$ Topic 2 (-11.9)
    \\ \midrule
    insurers\\
    prudential\\
    cuomo\\
    military\\
    benefits\\
    accounts\\
    richard\\
    metlife\\
    insurance\\
    yields \\
  \end{tabular}
&
  \begin{tabular}{c}
    $-$ Topic 3 (-10.8)
    \\ \midrule
    include\\
    closed\\
    whose\\
    mln\\
    range\\
    nasdaq\\
    raise\\
    among\\
    between\\
    trades \\ 
  \end{tabular}
\\ \midrule
  \begin{tabular}{c}
    $-$ Topic 4 (-10.6)
    \\ \midrule
    july\\
    nvidia\\
    exxon\\
    symantec\\
    motor\\
    semiconductor\\
    cut\\
    unemployment\\
    japan\\
    largest \\ 
  \end{tabular}
&
  \begin{tabular}{c}
    $-$ Topic 5 (-10.3)
    \\ \midrule
    goldman\\
    sachs\\
    buy\\
    rates\\
    list\\
    stocks\\
    prices\\
    adjusted\\
    demand\\
    result \\ 
  \end{tabular}
&
\\ \bottomrule
\end{tabular}
\end{table}

The sHDP learns strong topics that are assigned to fewer words and
indicate trends and tendencies at a finer-grained level, for example, on the level of actors instead of genres. The sHDP model is useful when more specific trends or tendencies are sought and when there is a possibility of overfitting or underfitting due to the number of topics.

\section{Conclusions}
We have presented a supervised Bayesian nonparametric model that
handles grouped data. Each group of data has an associated response
such as sentiment ratings or document popularity. The supervised HDP
(sHDP) model learns latent topics that are predictive of document
responses without having to choose a fixed number of topics, a
deficiency in previous models such as supervised LDA (sLDA)\@. In those
models, overfitting or underfitting can occur if the number of topics
is unsuitable for the dataset. The strongest topics learnt in the sHDP
are relatively finer-grained and are associated with fewer topics allowing
their effect on the document response to be learnt easily. Regression
and classification experiments were performed on real-world datasets
and showed that the model performs better than sLDA on
the newswire dataset, and only doing worse than sLDA on the movie snippet classification dataset. The experiments also showed that
jointly learning the topics and the GLM model produces topics and
results that are better than the simple alternative of learning the
topics unsupervised in a HDP model and training a regression model on top. Inference in the sHDP remains simple and is an adaptation of
that used in the HDP. The flexibility and ease of inference of the
sHDP means it has potential uses in many applications. Other inference techniques to improve performance can be explored such as variational inference \citep{Asuncion2009}. While the sHDP does not explicitly handle categorical outcomes, extra regression parameters for each topic can be added to do so.

While sentiment analysis models such as \citet{Pang2005} have a
similar goal of predicting document labels, the models we propose in
this paper are more general than typical sentiment analysis models and
do not require any bootstrap dictionary or labels for the terms. Our
models can additionally deal with a wide range of document response
types through a generalised linear model and can easily incorporate
additional information into its generative process as well as use
unlabelled data. The models in this
paper are not restricted to textual datasets as they can be used on
other kinds of data. For example, topic models have previously been
used on extracted image patches or image features by treating the
patches or features as words selected from a dictionary of
patches \citep{wang_image_classification}. Similarly, the models in this paper can be used to predict the keywords of an image or the theme of an image.

\ifCLASSOPTIONcaptionsoff
  \newpage
\fi

\bibliographystyle{IEEEtranN}
\bibliography{BiblioJab}

\begin{IEEEbiography}[{\includegraphics[width=1in,height=1.25in,clip,keepaspectratio]{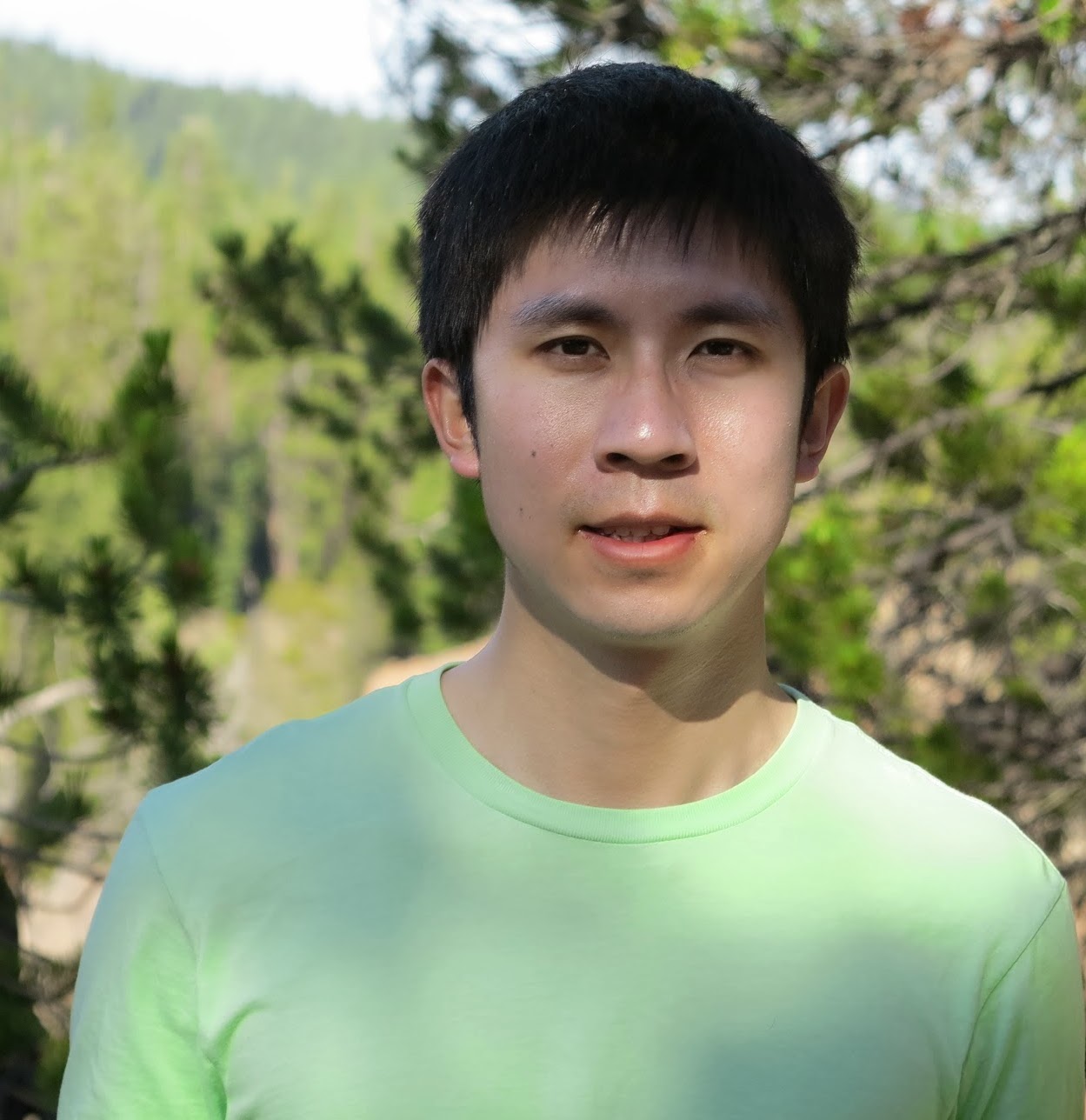}}]{Andrew M. Dai}
is a software engineer at Google Inc., Mountain View. He
completed this work during his PhD studies at the School of
Informatics, University of Edinburgh, and has a MA degree in computer
science from the University of Cambridge. His research interests include Bayesian
nonparametrics, topic modelling, Dirichlet processes, machine learning and collaborative filtering.
\end{IEEEbiography}

\begin{IEEEbiography}[{\includegraphics[width=1in,height=1.25in,clip,keepaspectratio]{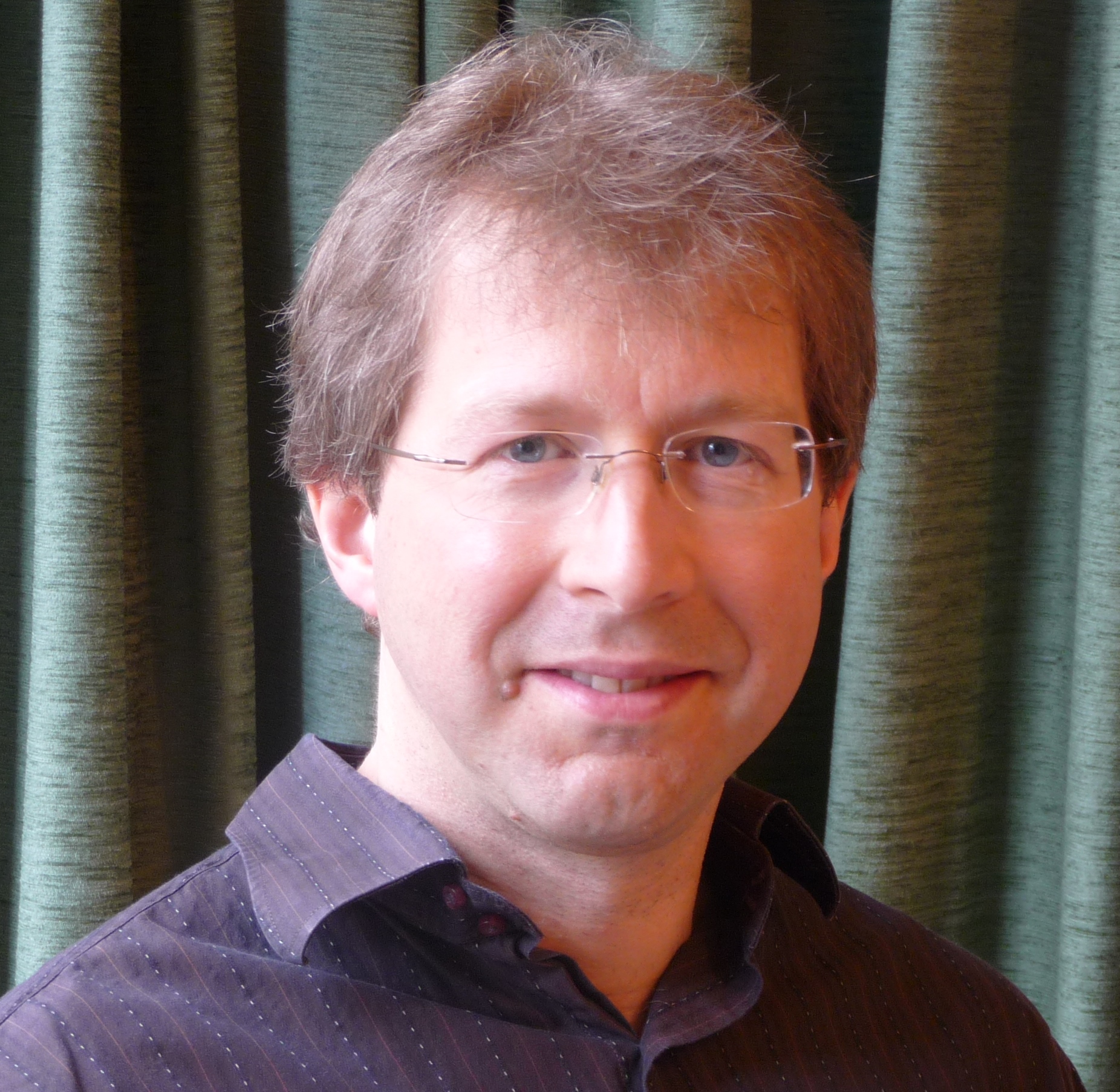}}]{Amos Storkey}
is a Reader in the School of Informatics, University of Edinburgh, with a background in mathematics (MA Maths Trinity, Cambridge), theoretical physics (Part III Maths), before focusing on Machine Learning (MSc, PhD Bayesian Gaussian Processes and Neural Networks, Imperial London). Prior to his current post he was research associate in the School of Informatics and Institute for Astronomy, Edinburgh and a Microsoft Research funded fellowship in Informatics, Edinburgh.  Storkey's research focuses on methods for incentivised distributed machine learning (e.g. Machine Learning Markets), inference and learning in continuous time systems, and applications in imaging and medical informatics. He is an associate editor of IEEE PAMI.
\end{IEEEbiography}

\end{document}